\documentclass[a4paper]{article}
\usepackage[pdftex]{graphicx}
\usepackage{multirow}
\usepackage{amssymb,amsmath,bm}
\newcommand{\argmax}{\mathop{\rm arg~max}\limits}
\usepackage{cite}
\usepackage{cases}
\usepackage{nidanfloat}
\usepackage{multirow}
\usepackage{comment}

\newcommand{\vsn}{\vspace{0pt}}

\newcommand{\vsntmp}{\vspace{0pt}}
\newcommand{\vsntmptmp}{\vspace{-4pt}}
\newcommand{\vsntmptmptmp}{\vspace{-1pt}}
\newcommand{\pvsntmptmptmp}{\vspace{1pt}}

\usepackage{INTERSPEECH2021}
\title{Cross-Modal Transformer-Based Neural Correction Models\\for Automatic Speech Recognition}
\name{}
\name{\begin{tabular}{c}
    Tomohiro Tanaka, Ryo Masumura, Mana Ihori, Akihiko Takashima,\\
    Takafumi Moriya, Takanori Ashihara, Shota Orihashi, Naoki Makishima\end{tabular}}
\address{NTT Media Intelligence Laboratories, NTT Corporation, Japan}
\email{tomohiro.tanaka.ht@hco.ntt.co.jp}
\begin{document}
\maketitle
\begin{abstract}
  We propose a cross-modal transformer-based neural correction models that refines the output of an automatic speech recognition (ASR) system so as to exclude ASR errors.
  Generally, neural correction models are composed of encoder-decoder networks, which can directly model sequence-to-sequence mapping problems.
  The most successful method is to use both input speech and its ASR output text as the input contexts for the encoder-decoder networks.
  However, the conventional method cannot take into account the relationships between these two different modal inputs because the input contexts are separately encoded for each modal.
  To effectively leverage the correlated information between the two different modal inputs, our proposed models encode two different contexts jointly on the basis of cross-modal self-attention using a transformer.
  We expect that cross-modal self-attention can effectively capture the relationships between two different modals for refining ASR hypotheses.
  We also introduce a shallow fusion technique to efficiently integrate the first-pass ASR model and our proposed neural correction model.
  Experiments on Japanese natural language ASR tasks demonstrated that our proposed models achieve better ASR performance than conventional neural correction models.
\end{abstract}
\section{Introduction}
\label{sec:intro}
Neural networks have dramatically improved the performance of automatic speech recognition (ASR) systems.
In particular, end-to-end ASR systems that directly convert an input speech utterance into an output text (e.g., characters, subwords, words) have attracted significant attention.
Recent studies have investigated various end-to-end ASR methods including
connectionist temporal classification \cite{DBLP:conf/icml/GravesJ14,DBLP:journals/corr/HannunCCCDEPSSCN14,DBLP:conf/asru/MiaoGM15},
recurrent neural network transducers \cite{graves_transd13,transd_asru17},
and attention-based encoder-decoder models with recurrent neural networks\cite{Chorowski_nips15,DBLP:conf/icassp/BahdanauCSBB16,DBLP:conf/interspeech/ZeyerISN18}
or transformers \cite{transf_asr_is18,all_you_need}.
Though these end-to-end ASR methods have made great progress,
their performance is insufficient when executing one-pass decoding using a single ASR model.

Various post-processing methods have been proposed to compensate for this insufficiency.
One existing methods is to use neural language models with fusion techniques \cite{shallow_fusion15,is17_shallow,cold_fusion18} or
rescoring hypotheses (or lattices) from first-pass ASR \cite{Mikolov2010rnnlm,Sundermeyer2012lstmlm,irie_is19_transf}.
Furthermore, methods that directly map ASR hypotheses to its reference text by using neural correction models have been investigated. 
One type of the neural correction models is text-to-text conversion-based models that convert the ASR hypothesis into the correct sentence \cite{is19_speller,icassp19_speller,compling19_text_norm,is19_trans_spell}.
Besides, the neural correction model in \cite{icassp20_delib} leverages both input speech and its ASR hypotheses as the input contexts for the encoder-decoder networks.
The model separately encodes the speech and hypothesis embeddings and individually attends to each encoded information with attention mechanisms \cite{Bahdanau_arxiv2014,NIPS2014_5346}.
However, this method cannot explicitly take into account where to correct the input ASR hypothesis because an encoder cannot mutually detect the positions of input in the other encoder.

We propose cross-modal transformer-based neural correction models that jointly encode speech and the ASR hypothesis to take their relationships into consideration.
Our proposed models use the self-attention in the transformer \cite{all_you_need} to take into account the cross-modal relationships between the speech and ASR hypothesis.
In our proposed models, two different modals are jointly encoded using cross-modal self-attention. 
We assume that the cross-modal self-attention enables the models to effectively learn the relationship between speech and hypothesis.
In contrast, a conventional neural correction model \cite{icassp20_delib} separately encodes the speech and ASR hypothesis.
In short, the advantage of our proposed models is that different modal inputs are effectively encoded while considering each other.
Our proposed models are assumed to explicitly take into account where to correct the input ASR hypothesis by capturing the relationship between it and the speech.
Finally, the decoder network computes the generative probabilities of tokens using the context vectors and generates the refined result by beam-search decoding.
Hu et al. used only the decoder of their neural correction model during beam-search decoding to generate refined results \cite{icassp20_delib}.
In contrast, we introduce shallow fusion \cite{shallow_fusion15,is17_shallow} to integrate first-pass ASR and neural correction models.
This enables us to complementarily use information from both decoders in ASR and neural correction models.
To the best of our knowledge, this is the first study that introduces cross-modal processing and shallow fusion into neural correction models for ASR.

We verify the effectiveness of our models with Japanese ASR tasks.
We prepare separate attention-based models to compare with our models.
\section{Related Work}
Neural correction models are closely related to neural language models.
In particular, encoder-decoder-based neural language models conditioned by speech information \cite{ttanaka_apsipa18,ttanaka_is18,icassp20_audiolm} and hypotheses from ASR \cite{tt_is18_neclm} are similar to neural correction models.
These language models were often applied so as to rescore ASR hypotheses, i.e., n-bests.
In contrast, neural correction models directly generate a sentence from ASR hypotheses and speech.

Our proposed models are related to studies that use self-attention to joinly encode cross-modal input features \cite{video_bert,DBLP:conf/cvpr/YeR0W19,DBLP:conf/acl/ThapliyalS20}.
These studies proposed cross-modal attention of visual and linguistic features with transformer-based neural networks.
In this paper, we focus on speech and text information for neural correction models for ASR and use self-attention to take into account the relationships between them.
\section{End-to-End ASR}
An end-to-end ASR system directly converts acoustic features of input speech into a token sequence.
In this study, we use auto-regressive encoder-decoder models that predict the probability of a token given the previous predicted tokens.
Given speech $\bm{X} = \{\bm{x}_1, \cdots ,\bm{x}_I\}$, the encoder-decoder estimates the generative probability of a token sequence $\bm{W} = \{w_1, \cdots ,w_T\}$,
where $I$ is the number of the acoustic features in the input speech and $T$ is the number of the tokens in the token sequence.
The generative probability of a token sequence is defined as
\begin{align}
  P(\bm{W}|\bm{X}; \bm{\Lambda}) = \prod_{t=1}^T P(w_t| w_{1:t-1}, \bm{X};\bm{\Lambda}),
  \label{eq:prob_r}
\end{align}
where $\bm{\Lambda}$ represents the trainable parameters.

The model parameters $\bm{\Lambda}$ in an end-to-end ASR system are updated to maximize the generative probability in the decoder when give an input speech.
Thus, the model parameters are optimized by minimizing the cross entropy loss function:
\begin{equation}
  \mathcal{L}(\bm{\Lambda}) = - \sum_{(\bm{W}^{\prime}, \bm{X}^{\prime}) \in \mathcal{D}} \log P(\bm{W}^{\prime}|\bm{X}^{\prime}; \bm{\Lambda}),
\end{equation}
where $\mathcal{D}$ is the training set.

\section{Transformer-Based Neural Correction Models for ASR}
An ASR system generates a hypothesis from an input speech.
Neural correction models then generate a refined result from the hypothesis and input speech.
We use a transformer encoder-decoder for neural correction models.
Given the acoustic feature sequence $\bm{X} = \{\bm{x}_1, \cdots ,\bm{x}_I\}$ and the ASR output generated by the ASR system $F(\cdot)$,
neural correction models estimate the generative probability of $\bm{W} = \{w_1, \cdots ,w_T\}$ as
\begin{align}
P(\bm{W}| \bm{X}; \bm{\Theta}, \bm{\Lambda}) = & \prod_{t=1}^T P(w_t|w_{1:t-1}, \bm{X}, F(\bm{X}; \bm{\Lambda}); \bm{\Theta}) \\
= & \prod_{t=1}^T P(w_t|w_{1:t-1}, \bm{X}, \bm{C}; \bm{\Theta}),
\label{eq:ncm}
\end{align}
where $\bm{C}=\{ c_1, \cdots ,c_J\}$ is the ASR hypothesis from the ASR system
and $\bm{\Theta}$ represents the model parameters in the neural correction model.
\subsection{Proposed Method}
\label{sec:cross}
\begin{figure}[t]
  \centering
  \includegraphics[width=0.7 \linewidth]{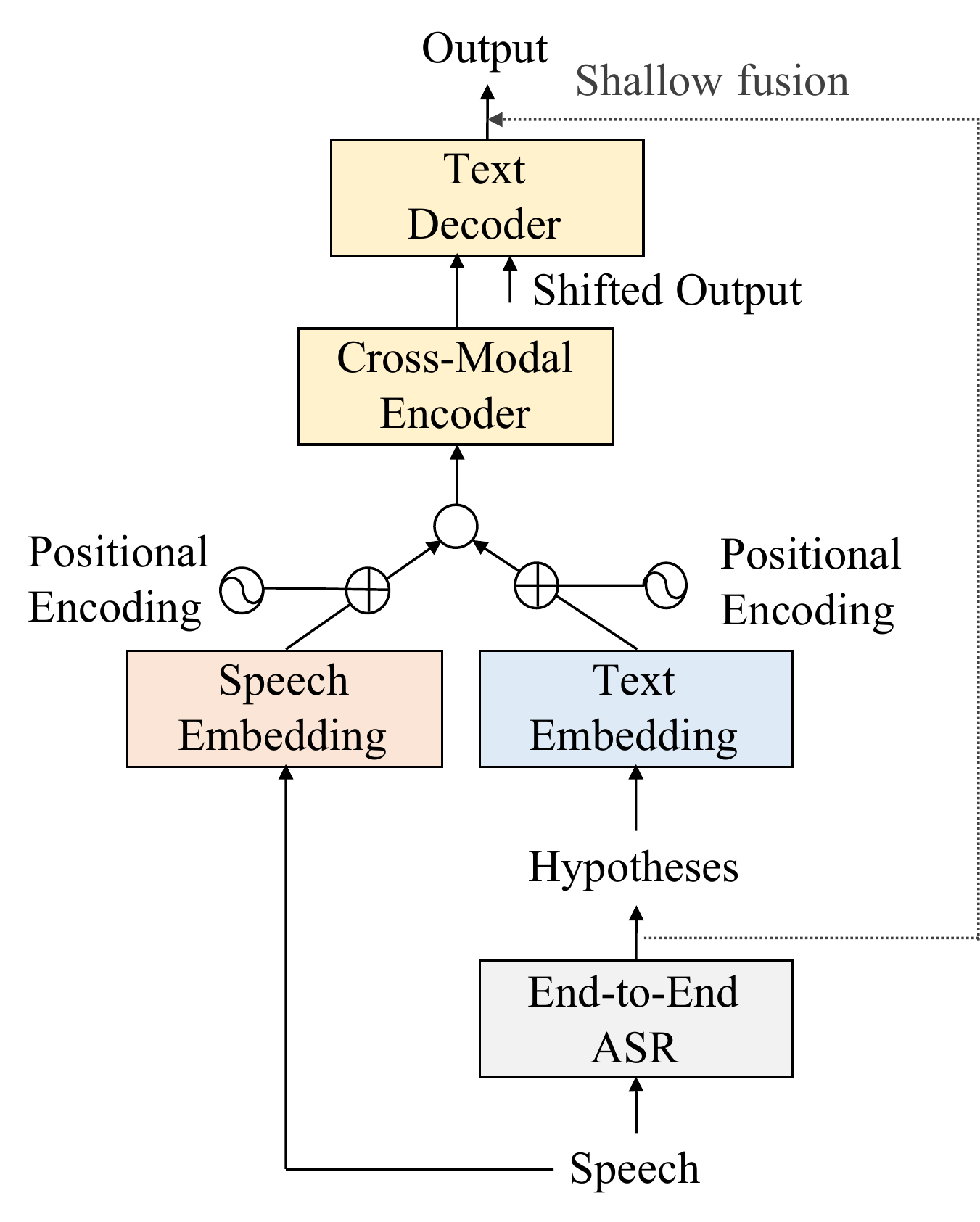}
  \pvsntmptmptmp
  \caption{Network structure of proposed cross-modal transformer-based neural correction model.}
  \label{fig:cross}
  \pvsntmptmptmp
\end{figure}
Figure \ref{fig:cross} illustrates the network structure of our proposed models that involves encoding speech and the ASR hypothesis with a cross-modal transformer.
Our proposed models consist of embedding networks for speech and hypothesis.
\smallskip
\\{\bf Speech Embedding: }The acoustic features are embedded into continuous representations as
\begin{align}
  \overline{\bm{x}}_{i} & = \mathtt{SpeechEmbedding}(\bm{x}_i, {\theta}_x), \label{eq:spemb1} \\
  \bm{s}_{i} & = \mathtt{PosEncoding}(\overline{\bm{x}}_{i}), \label{eq:spemb2}
\end{align}
where $\mathtt{SpeechEmbedding}(\cdot)$ is a neural network that converts acoustic features into continuous vectors,
$\mathtt{PosEncoding}(\cdot)$ is a function that adds a continuous vector in which position information is embedded 
and ${\theta}_x$ is the trainable parameter.
\smallskip
\\{\bf Text Embedding: }Each token $c_j$ in a hypothesis $\bm{C} = \{c_1, \cdots,c_J\}$ is encoded to one-hot representation and embedded into continuous representation as
\begin{align}
  \overline{\bm{c}}_{j} & = \mathtt{TextEmbedding}(c_j; {\theta}_d), \label{eq:hypemb1} \\
  \bm{d}_{j} & = \mathtt{PosEncoding}(\overline{\bm{c}}_{j}), \label{eq:hypemb2}
\end{align}
where $\mathtt{TextEmbedding}(\cdot)$ is a function that converts a token into a continuous representation,
and ${\theta}_d$ is a trainable parameter.
\smallskip
\\{\bf Cross-Modal Encoder: } The two embedded vectors are concatenated into a single sequence $\bm{e}$ as
\begin{align}
  \bm{e} = \{ \bm{s}_1, \ldots, \bm{s}_I, \bm{s}_{\rm sep} ,\bm{d}_1, \ldots, \bm{d}_J \} ,
\label{eq:concat}
\end{align}
where $\bm{s}_{\rm sep}$ is a continuous representation of a separator token ${\tt [sep]}$.
We define the output of the $m$-th transformer encoder block as $\bm{f}^{m}$ and the input of the first block is defined as $\bm{f}^{0} = \bm{e}$.
The computational process in the $m$-th block of the transformer encoder is defined as
\begin{align}
  \bm{f}^m & = \mathtt{CrossModalEncoder}(\bm{f}^{m-1}; {\theta}_f),
  \label{eq:trenc}
\end{align}
where $\mathtt{CrossModalEncoder}(\cdot)$ is a transformer encoder including a scaled dot-product multi-head self-attention layer and a position-wise feed-forward network
and ${\theta}_f$ is a trainable parameter.
\smallskip
\\{\bf Text Decoder: }The output of the final block $\bm{f}^M$ in the cross-modal encoder is fed into the transformer decoder.
When the output of the $t$-th time step for the $n$-th transformer block in the decoder is $\bm{q}_{t}^{n-1}$,
the transformer decoder constructs a hidden representation with the context vector $\bm{U}_{t}^{n}$ from the encoder as
\begin{align}
  \bm{U}_{t}^{n} & =\mathtt{SrcTgtAttention}(\bm{f}^M, \bm{q}_{t-1}^{n-1} ;{\theta}_U),\\
  \bm{q}_{t}^{n} & = \mathtt{TransformerDecoder}(\bm{q}_{1:t-1}^{n-1}, \bm{U}_{t}^{n}; {\theta}_q),
  \label{eq:trdec}
\end{align}
where $\mathtt{TransformerDecoder}(\cdot)$ is a transformer decoder including a scaled dot-product multi-head masked self-attention layer and a position-wise feed-forward network,
$\mathtt{SrcTgtAttention}(\cdot)$ is a scaled dot product multi-head source-target attention layer,
$M$ is the number of blocks in the transformer encoder,
and ${\theta}_h$ is the trainable parameter.
The input of the first block is token embedding, which is calculated as
\begin{align}
   \overline{\bm{w}}_{t} & = \mathtt{TextEmbedding}(w_t; {\theta}_w),\\
   \bm{q}^{0}_{t} & = \mathtt{PosEncoding}(\overline{\bm{w}}_{t}),
\end{align}
where ${\theta}_w$ is the trainable parameter.
The network estimates the probabilities of a distribution of the output tokens as
\begin{align}
  P(w_t|w_{1:t-1}, \bm{X}, \bm{C}; \bm{\Theta}) = \mathtt{Softmax}(\bm{q}_{1:t-1}^{N}; {\theta}_o),
  \label{eq:softmax}
\end{align}
where $\mathtt{Softmax}(\cdot)$ represents the softmax function with linear transformation,
$N$ is the number of blocks in the transformer decoder,
and ${\theta}_o$ is the trainable parameter.
Finally, beam search decoding is conducted while calculating the probability distribution.
The model parameters can be summarized as $\bm{\Theta} = \{ \theta_x,\theta_d,\theta_f,\theta_U,\theta_q,\theta_w,\theta_o \}$.
\subsection{Conventional Method}
\label{sec:separate}
\begin{figure}[t]
  \centering
  \includegraphics[width=0.7 \linewidth]{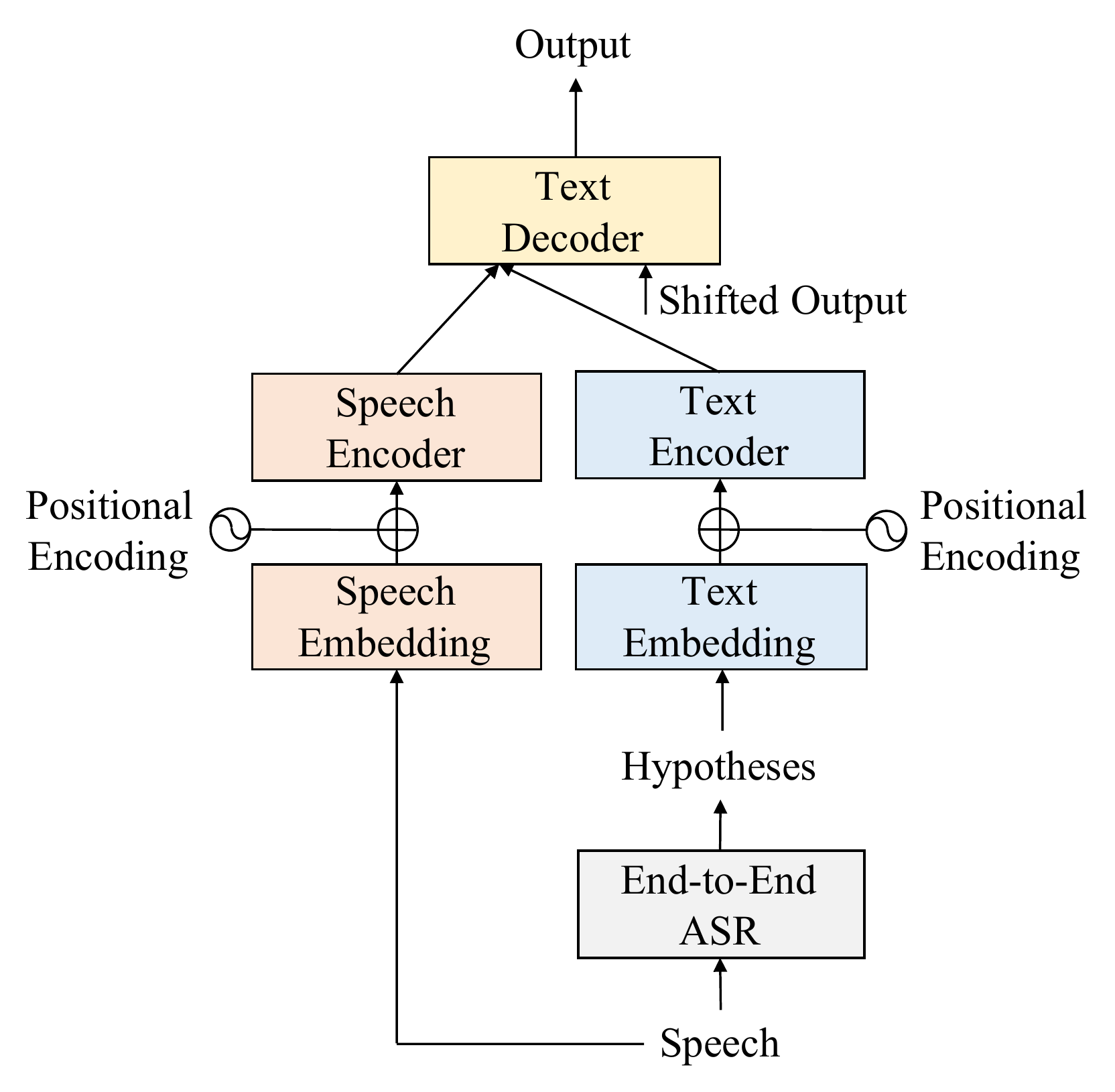}
  \caption{Network structure of conventional separate attention-based neural correction model.}
  \label{fig:separate}
  \vsntmptmptmp
\end{figure}
Figure \ref{fig:separate} illustrates the network structure of a conventional neural correction model that separately encodes speech and hypothesis by each encoder.
The input speech and hypothesis are converted into continuous representations $\bm{s}_{i}$ and $\bm{d}_j$ in the same manner as in Eqs. (\ref{eq:spemb1}--\ref{eq:spemb2}) and (\ref{eq:hypemb1}--\ref{eq:hypemb2}).
The embeddings $\bm{s}_{1:I}$ and $\bm{d}_{1:J}$ are input into the speech and text encoders.

We define the output of the $l$-th speech encoder block and $m$-th text encoder block as $\bm{f}^{l}$ and $\bm{g}^{m}$ respectively.
In this case, the inputs of the first blocks are defined as $\bm{f}^{0} = \bm{s}_{1:I}$ and $\bm{g}^{0} = \bm{d}_{1:J}$.
The computational process in the $l$-th block of the transformer speech encoder and $m$-th block of the transformer text encoder are defined as
\begin{align}
  \bm{f}^{l} & = \mathtt{SpeechEncoder}(\bm{f}^{l-1}; {\theta}_f),\\
  \bm{g}^{m} & = \mathtt{TextEncoder}(\bm{g}^{m-1}; {\theta}_g),
  \label{eq:sp-txtenc}
\end{align}
where $\mathtt{SpeechEncoder}(\cdot)$ and $\mathtt{TextEncoder}(\cdot)$ are transformer encoders including a scaled dot product multi-head self-attention layer and a position-wise feed-forward network and 
${\theta}_f$ and ${\theta}_g$ are the trainable parameters.
The final outputs of the two encoders are attended to by source-target attention mechanisms and concatenated as
\begin{align}
  \overline{\bm{U}}_{t}^{n} & =\mathtt{SrcTgtAttention}(\bm{f}^L, \bm{q}_{t-1}^{n-1} ;{\theta}_{\overline{U}}),\\
  \overline{\overline{\bm{U}}}_{t}^{n} & =\mathtt{SrcTgtAttention}(\bm{g}^M, \bm{q}_{t-1}^{n-1} ;{\theta}_{\overline{\overline{U}}}),\\
  \bm{U}_{t}^{n} & = [{\overline{\bm{U}}_{t}^{n}}^T, {\overline{\overline{\bm{U}}}_{t}^{n}}^T ]^T,
\end{align}
where $L$ is the number of blocks in the transformer of the speech encoder, and $M$ is the number of blocks in the transformer of the text encoder.
The decoder and the probability calculation are the same as those in the proposed model (Eq.(\ref{eq:trdec}–\ref{eq:softmax})).
The model parameters can be summarized as $\bm{\Theta} = \{ \theta_x,\theta_d,\theta_f,\theta_g, \theta_{\overline{U}}, \theta_{\overline{\overline{U}}}, \theta_q,\theta_w,\theta_o \}$.
\subsection{Shallow Fusion of ASR and Neural Correction Model }
Shallow fusion \cite{shallow_fusion15,is17_shallow} is a technique to incorporate an external model by log-linear interpolation at inference time.
We use this technique to merge the first-pass ASR model and neural correction model.
We use the following criterion during beam search decoding:
\begin{align}
  \hat{\bm{W}} = \argmax_{\bm{W}} (1 - \alpha) & \log P(\bm{W}|\bm{X}; \bm{\Lambda}) + \nonumber
  \\&  \alpha \log P(\bm{W}|\bm{X}; \bm{\Lambda}, \bm{\Theta}),
\end{align}
where $\alpha$ is the weight of the neural correction model.
\subsection{Training}
In the training step, the parameters $\bm{\Theta}$ for a neural correction model are updated to maximize the conditional generative probability in the decoder
when given an input speech and the ASR hypothesis as a context from the encoder.
Thus, these parameters are optimized by minimizing the cross entropy loss function:
\begin{equation}
  \mathcal{L}(\bm{\Theta}) = - \sum_{(\bm{W}^{\prime}, \bm{X}^{\prime}, \bm{C}^{\prime}) \in \mathcal{D}^{\prime}} \log P(\bm{W}^{\prime}|\bm{X}^{\prime}, \bm{C}^{\prime}; \bm{\Theta}),
\end{equation}
where $\mathcal{D}^{\prime}$ is the training set.
Our proposed and conventional models can be trained with the same criterion.
Note that these neural correction models can be trained from the same training data as ASR since the $\bm{C}$ is obtained from function $F(\cdot)$ in Eq. (\ref{eq:ncm}).
\section{Experiments}
\vsntmptmp
\label{sec:exp}
\subsection{Setups}
We used two corpora: the corpus of spontaneous Japanese (CSJ) \cite{CSJ} and our home-made corpus of natural two-person dialogue corpus (NTDC), which consists of about 24 hours of speech and its transcriptions.
We split NTDC into 22 hours for the training set, 1 hour for development set, and 1 hour for the evaluation set.
CSJ, which has about 545 hours of speech, was used for training data.
We used three standard evaluation sets for CSJ (CSJ1, CSJ2, and CSJ3).

We used a transformer-based encoder-decoder ASR model as the baseline (\textbf{Baseline}).
The encoder and decoder each had six transformer blocks.
The token embedding dimension, hidden state dimension, non-linear layer dimension, and the number of heads were 256, 256, 2048, and 4, respectively.
The acoustic features were transformed by two layers of 2D convolutional neural network.
In our proposed cross-modal transformer-based neural correction model (\textbf{Cross-Modal}), the speech embedding also constructed using a two-layer 2D CNN.
The text embeddings were 256-dimensional continuous representations.
The transformer encoder and decoder had six blocks each.
The hidden state dimension, non-linear layer dimension, and the number of heads are 256, 2048, and 4, respectively.
The conventional model was the neural correction model with separate attention (\textbf{Separate}), which had an encoder for speech and one for text.
The conventional neural correction model had two encoders for speech and text respectively.
The encoders for speech and text each had six transformer blocks respectively.
The other configurations of the transformer had the same values as those of \textbf{Cross-Modal}.
We applied shallow fusion (\textbf{SF}) to both models for comparison.
\begin{table}[t]
  \centering
  \vsntmptmp
\caption{CERs (\%) on evaluation sets of each data set. SF denotes shallow fusion of baseline and neural correction model.}
\vsntmptmp\vsntmptmp
\label{tab:main}
\begin{tabular}{l|rrrr|r}
\hline
\multicolumn{1}{l|}{Model}     & NTDC        & CSJ1       & CSJ2       & CSJ3        & All  \\ \hline
Baseline                       & 20.3        & 10.1       & 6.5        & 9.4         & 10.5 \\
Separate                       & 20.4        & 10.0       & 6.5        & 8.9         & 10.4 \\
\quad+SF                       & 20.4        & 9.9        & $\bm{6.4}$ & 8.9         & 10.3 \\
Cross-Modal                    & 20.3        & 10.0       & $\bm{6.4}$ & 9.1         & 10.4 \\
\quad+SF                       & $\bm{20.2}$ & $\bm{9.4}$ & $\bm{6.4}$ &  $\bm{8.4}$ & $\bm{10.0}$ \\ \hline
\end{tabular}
\vsn
\vsntmptmptmp
\end{table}
\setcounter{figure}{3}
\begin{figure*}[t]
  \centering
  \includegraphics[width=0.94\linewidth]{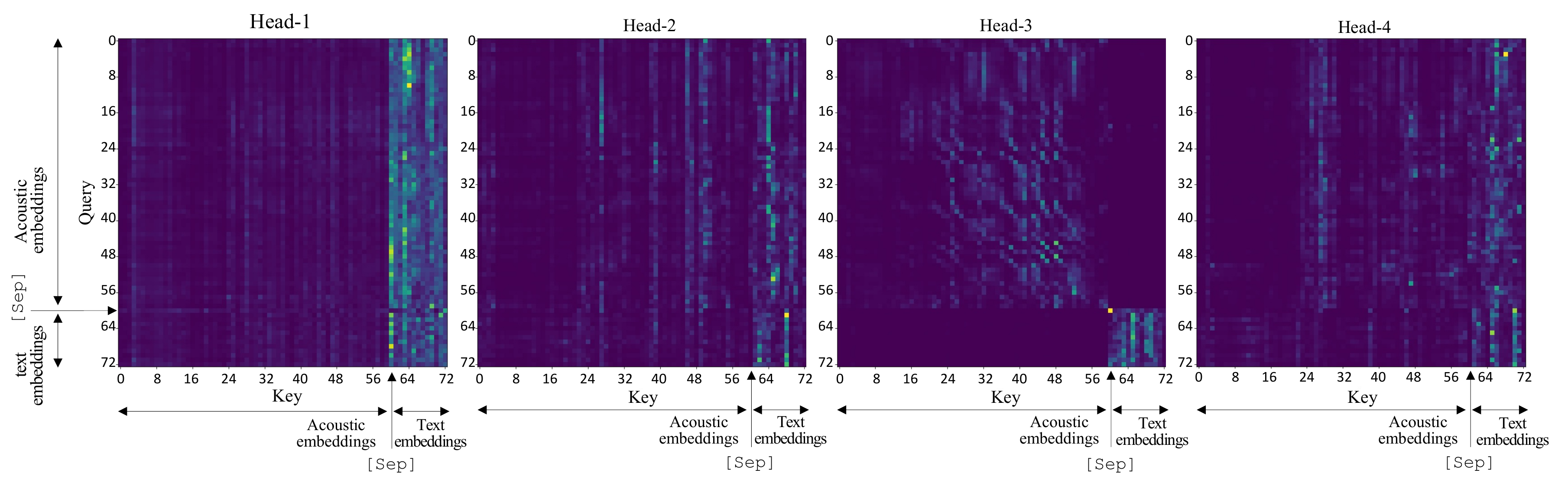}
  \vsntmptmp\vsntmptmp
  \caption{Sample of self-attention weights in cross-modal transformer. Number of images is equal to the number of heads in the transformer encoder. ${\tt [sep]}$ is a separator token.}
  \label{fig:vis}
  \vsn
  \vsntmp
\end{figure*}

The acoustic features was a 40-dimensional log Mel-filterbank with delta and acceleration coefficients.
We applied SpecAugument\cite{spaug} during training of all models.
The vocabulary size was 3285 characters made from CSJ and NTDC.
In the cross-modal transformer-based model, we added the special token ${\tt [sep]}$ to the vocabulary.
We used Adam optimizer with Noam learning rate scheduler with 25000 warmup steps.
When decoding by beam search, the beam size was set to 20.
\subsection{Results}
\label{sec:res}
Table \ref{tab:main} shows the character error rate (CER) performance when using the baseline system and two correction-based systems with and without shallow fusion.
Cross-modal transformer-based model and separate attention-based model showed similar performance when shallow fusion was not performed.
When using shallow fusion, the cross-modal transformer-based model outperformed the separate attention-based model.
These results indicated that cross-modal transformer effectively learned the relationship between the speech and hypothesis.
We found that shallow fusion efficiently integrated ASR and cross-modal transformer-based model.

Figure \ref{fig:weight} shows the relationship between the weights of neural correction models and CER in shallow fusion.
Cross-modal transformer-based model showed better performance over a wider range than the separate attention-based model.
Cross-modal transformer-based and separate attention-based model with shallow fusion showed better performance than baseline for all weights.
This indicates that shallow fusion can efficiently integrate ASR and neural correction models.

Figure \ref{fig:vis} gives an example of the self-attention weights between key and query values in the first block for the cross-modal transformer-based model.
Each image shows the weights of each head in the self-attention in the transformer encoder.
The network detected the boundaries and extracted the features of both speech and ASR-hypothesis embeddings.
It is assumed that head-1 extracted ASR-hypothesis information to determine the correspondence with the ASR hypothesis and output text,
head-2 and head-4 extracted both speech and ASR-hypothesis information to determine the correspondence with the two embeddings, and
head-3 extracted speech information to determine the correspondence with speech and output text.
All the images in Figure \ref{fig:vis} show that the cross-modal transformer functioned as expected.
\section{Conclusions}
\label{sec:conc}
\setcounter{figure}{2}
\begin{figure}[t]
  \vsntmptmptmp\vsntmptmptmp\vsntmptmptmp\vsntmptmptmp\vsntmptmptmp\vsntmptmptmp\vsntmptmptmp\vsntmptmptmp\vsntmptmptmp\vsntmptmptmp\vsntmptmptmp\vsntmptmptmp
  \vsntmptmp\vsntmptmp\vsntmptmp\vsntmptmp
  \centering
  \includegraphics[width=0.94\linewidth]{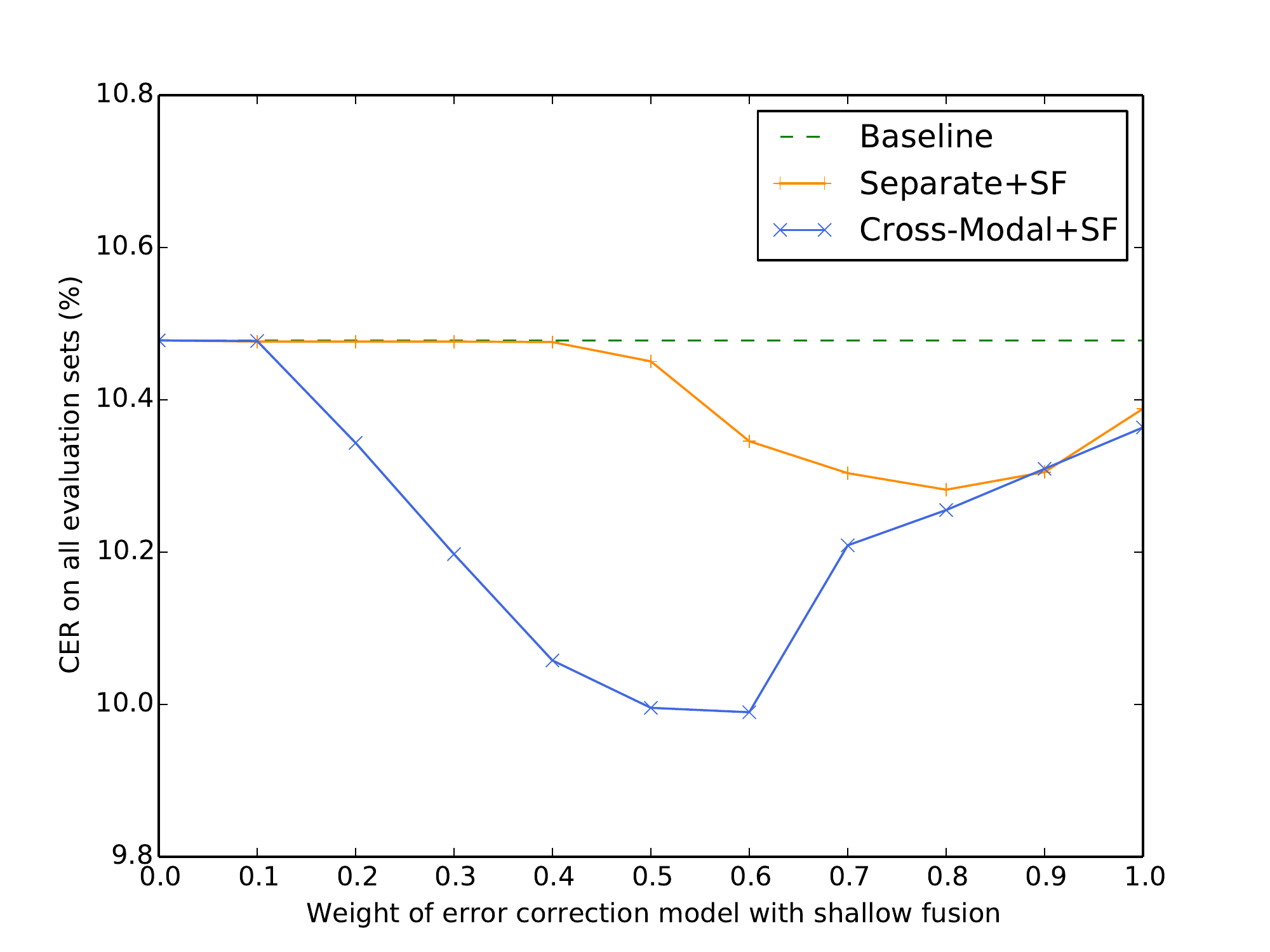}
  \vsntmptmp\vsntmptmp\vsntmptmp
  \caption{CERs on all evaluation sets in different weights of neural correction model with shallow fusion.}
  \label{fig:weight}
  \vsntmptmp\vsntmptmp\vsntmptmptmp
  \vsntmptmptmp
\end{figure}
We proposed cross-modal transformer-based neural correction models, which jointly encode speech and hypothesis.
The key strength of our proposed models is effectively capturing the relationships between two different modals for refining ASR hypotheses.
From experiments involving CERs, our model with shallow fusion showed the best performance of all the models in our experiments.

\clearpage

\begin{thebibliography}{10}

  \bibitem{DBLP:conf/icml/GravesJ14}
    A.~Graves and N.~Jaitly, ``Towards end-to-end speech recognition with recurrent neural networks,'' \emph{Proc. International Conference on Machine Learning, ({ICML})}, pp. 1764--1772, 2014.
    
  \bibitem{DBLP:journals/corr/HannunCCCDEPSSCN14}
    A.~Y. Hannun, C.~Case, J.~Casper, B.~Catanzaro, G.~Diamos, E.~Elsen,
    R.~Prenger, S.~Satheesh, S.~Sengupta, A.~Coates, and A.~Y. Ng, ``Deep speech:
    Scaling up end-to-end speech recognition,'' \emph{arXiv: 1412.5567}, 2014.
    
  \bibitem{DBLP:conf/asru/MiaoGM15}
    Y.~Miao, M.~Gowayyed, and F.~Metze, ``{EESEN:} end-to-end speech recognition
    using deep {RNN} models and wfst-based decoding,'' \emph{Proc. Workshop on
    Automatic Speech Recognition and Understanding (ASRU)}, pp. 167--174, 2015.
    
  \bibitem{graves_transd13}
    A.~Graves, A.~Mohamed, and G.~E. Hinton, ``Speech recognition with deep
    recurrent neural networks,'' \emph{In Proc. of International Conference on
    Acoustics, Speech and Signal Processing (ICASSP)}, pp. 6645--6649, 2013.
    
  \bibitem{transd_asru17}
    K.~Rao, H.~Sak, and R.~Prabhavalkar, ``Exploring architectures, data and units
    for streaming end-to-end speech recognition with rnn-transducer,'' \emph{Proc
    of Automatic Speech Recognition and Understanding Workshop (ASRU)}, pp.
    193--199, 2017.
    
  \bibitem{Chorowski_nips15}
    J.~Chorowski, D.~Bahdanau, D.~Serdyuk, K.~Cho, and Y.~Bengio, ``Attention-based
    models for speech recognition,'' \emph{In Proc. Annual Conference on Neural
    Information Processing Systems (NIPS)}, pp. 577--585, 2015.
    
  \bibitem{DBLP:conf/icassp/BahdanauCSBB16}
    D.~Bahdanau, J.~Chorowski, D.~Serdyuk, P.~Brakel, and Y.~Bengio, ``End-to-end
    attention-based large vocabulary speech recognition,'' \emph{Proc.
    International Conference on Acoustics, Speech and Signal Processing
    (ICASSP)}, pp. 4945--4949, 2016.
    
  \bibitem{DBLP:conf/interspeech/ZeyerISN18}
    A.~Zeyer, K.~Irie, R.~Schl{\"{u}}ter, and H.~Ney, ``Improved training of
    end-to-end attention models for speech recognition,'' \emph{Proc. of Annual
    Conference of the International Speech Communication Association
    (INTERSPEECH)}, pp. 7--11, 2018.
    
  \bibitem{transf_asr_is18}
    L.~Dong, S.~Xu, and B.~Xu, ``Speech-transformer: {A} no-recurrence
    sequence-to-sequence model for speech recognition,'' \emph{Proc. of
    International Conference on Acoustics, Speech and Signal Processing
    ({ICASSP})}, pp. 5884--5888, 2018.
    
  \bibitem{all_you_need}
    A.~Vaswani, N.~Shazeer, N.~Parmar, J.~Uszkoreit, L.~Jones, A.~N. Gomez,
    L.~Kaiser, and I.~Polosukhin, ``Attention is all you need,'' \emph{Annual
    Conference on Neural Information Processing Systems (NeurIPS)}, pp.
    5998--6008, 2017.
    
  \bibitem{shallow_fusion15}
    {\c{C}}.~G{\"{u}}l{\c{c}}ehre, O.~Firat, K.~Xu, K.~Cho, L.~Barrault, H.~Lin,
    F.~Bougares, H.~Schwenk, and Y.~Bengio, ``On using monolingual corpora in
    neural machine translation,'' \emph{CoRR}, vol. abs/1503.03535, 2015.
    
  \bibitem{is17_shallow}
    J.~Chorowski and N.~Jaitly, ``Towards better decoding and language model
    integration in sequence to sequence models,'' \emph{In Proc. of International
    Speech Communication Association (INTERSPEECH)}, pp. 523--527, 2017.
  
  \bibitem{cold_fusion18}
    A.~Sriram, H.~Jun, S.~Satheesh, and A.~Coates, ``Cold fusion: Training seq2seq
    models together with language models,'' \emph{Proc. of the Conference of the
    International Speech Communication Association (INTERSPEECH)}, pp. 387--391,
    2018.
  
  \bibitem{Mikolov2010rnnlm}
    T.~Mikolov, M.~Karafi{\'{a}}t, L.~Burget, J.~Cernock{\'{y}}, and S.~Khudanpur,
    ``Recurrent neural network based language model,'' \emph{Proc. Annual
    Conference of the International Speech Communication Association
    ({INTERSPEECH})}, pp. 1045--1048, 2010.
    
  \bibitem{Sundermeyer2012lstmlm}
    M.~Sundermeyer, R.~Schl{\"{u}}ter, and H.~Ney, ``{LSTM} neural networks for
    language modeling,'' \emph{Proc. Annual Conference of the International
    Speech Communication Association ({INTERSPEECH})}, pp. 194--197, 2012.
    
  \bibitem{irie_is19_transf}
    K.~Irie, A.~Zeyer, R.~Schl{\"{u}}ter, and H.~Ney, ``Language modeling with deep
    transformers,'' \emph{Proc. of International Speech Communication Association
    (INTERSPEECH)}, pp. 3905--3909, 2019.
    
  \bibitem{is19_speller}
    S.~Zhang, M.~Lei, and Z.~Yan, ``Investigation of transformer based spelling
    correction model for ctc-based end-to-end mandarin speech recognition,''
    \emph{Proc. of the Conference of the International Speech Communication
    Association (INTERSPEECH)}, pp. 2180--2184, 2019.

  \bibitem{icassp19_speller}
    J.~Guo, T.~N. Sainath, and R.~J. Weiss, ``A spelling correction model for
    end-to-end speech recognition,'' \emph{Proc. of the International Conference
    on Acoustics, Speech and Signal Processing (ICASSP)}, pp. 5651--5655, 2019.

  \bibitem{compling19_text_norm}
    H.~Zhang, R.~Sproat, A.~H. Ng, F.~Stahlberg, X.~Peng, K.~Gorman, and B.~Roark,
    ``Neural models of text normalization for speech applications,''
    \emph{Comput. Linguistics}, vol.~45, no.~2, pp. 293--337, 2019.

  \bibitem{is19_trans_spell}
    S.~Zhang, M.~Lei, and Z.~Yan, ``Investigation of transformer based spelling
    correction model for ctc-based end-to-end mandarin speech recognition,''
    \emph{Proc. the Conference of the International Speech Communication
    Association (INTERSPEECH)}, pp. 2180--2184, 2019.

  \bibitem{icassp20_delib}
    K.~Hu, T.~N. Sainath, R.~Pang, and R.~Prabhavalkar, ``Deliberation model based
    two-pass end-to-end speech recognition,'' \emph{Proc. of International
    Conference on Acoustics, Speech and Signal Processing ({ICASSP})}, pp.
    7799--7803, 2020.

  \bibitem{Bahdanau_arxiv2014}
    D.~Bahdanau, K.~Cho, and Y.~Bengio, ``Neural machine translation by jointly
    learning to align and translate,'' \emph{arXiv:1409.0473}, 2014.

  \bibitem{NIPS2014_5346}
    I.~Sutskever, O.~Vinyals, and Q.~V. Le, ``Sequence to sequence learning with
    neural networks,'' \emph{In Proc. Annual Conference on Neural Information
    Processing Systems (NIPS)}, pp. 3104--3112, 2014.

  \bibitem{ttanaka_apsipa18}
    T.~Tanaka, R.~Masumura, T.~Moriya, and Y.~Aono, ``Neural speech-to-text
    language models for rescoring hypotheses of {DNN-HMM} hybrid automatic speech
    recognition systems,'' \emph{Asia-Pacific Signal and Information Processing
    Association Annual Summit and Conference (APSIPA-ASC)}, pp. 196--200, 2018.

  \bibitem{ttanaka_is18}
    T.~Tanaka, R.~Masumura, T.~Moriya, T.~Oba, and Y.~Aono, ``A joint end-to-end
    and {DNN-HMM} hybrid automatic speech recognition system with transferring
    sharable knowledge,'' \emph{Proc. of the Conference of the International
    Speech Communication Association (INTERSPEECH)}, pp. 2210--2214, 2019.

  \bibitem{icassp20_audiolm}
    A.~Gandhe and A.~Rastrow, ``Audio-attention discriminative language model for
    {ASR} rescoring,'' \emph{Proc. of the International Conference on Acoustics,
    Speech and Signal Processing ({ICASSP})}, pp. 7944--7948, 2020.

  \bibitem{tt_is18_neclm}
    T.~Tanaka, R.~Masumura, H.~Masataki, and Y.~Aono, ``Neural error corrective
    language models for automatic speech recognition,'' \emph{Proc. of the
    International Speech Communication Association {INTESPEECH}}, pp. 401--405,
    2018.

  \bibitem{video_bert}
    C.~Sun, A.~Myers, C.~Vondrick, K.~Murphy, and C.~Schmid, ``Videobert: {A} joint
    model for video and language representation learning,'' \emph{Proc. of
    International Conference on Computer Vision (ICCV)}, pp. 7463--7472, 2019.
    
  \bibitem{DBLP:conf/cvpr/YeR0W19}
    L.~Ye, M.~Rochan, Z.~Liu, and Y.~Wang, ``Cross-modal self-attention network for
    referring image segmentation,'' \emph{Proc. of Conference on Computer Vision
    and Pattern Recognition ({CVPR})}, pp. 10\,502--10\,511, 2019.
    
  \bibitem{DBLP:conf/acl/ThapliyalS20}
    A.~V. Thapliyal and R.~Soricut, ``Cross-modal language generation using pivot
    stabilization for web-scale language coverage,'' \emph{Proc. of the Annual
    Meeting of the Association for Computational Linguistics ({ACL})}, pp.
    160--170, 2020.

  \bibitem{CSJ}
    S.~Furui, K.~Maekawa, and H.~Isahara, ``A {J}apanese national project on
    spontaneous speech corpus and processing technology,'' \emph{In Proc. ASR2000
    - Automatic Speech Recognition: Challenges for the new Millenium}, pp.
    244--248, 2000.

  \bibitem{spaug}
    D.~S. Park, W.~Chan, Y.~Zhang, C.~Chiu, B.~Zoph, E.~D. Cubuk, and Q.~V. Le,
    ``Specaugment: {A} simple data augmentation method for automatic speech
    recognition,'' \emph{In Proc. of Conference of the International Speech
    Communication Association (INTERSPEECH)}, pp. 2613--2617, 2019.
  
\end{thebibliography}

\end{document}